# Clause Encounters of the Third Kind: Can LLMs Replace Language Teachers?

**Kristina Šekrst**

University of Zagreb

ORCID: 0000-0002-0467-7313

**Ana Kovačić**

University of Zagreb

ORCID: 0000-0002-8414-5041

**Abstract**

While various organizations now actively encourage LLM use in classrooms, we still lack rigorous, systematic evaluations of how well these models actually perform the fundamental tasks of language pedagogy. This paper examines whether state-of-the-art LLMs can deliver the kind of corrective feedback and methodological explanations that language learners need. The study tests multiple large language models on their ability to identify, correct, and explain common learner mistakes in English, by systematically varying model parameters to investigate how these technical adjustments affect output quality, pedagogical clarity, and consistency, along with using retrieval-augmented generation to query methodological data. The evaluation employs automated metrics (GLEU, BERTScore) but also human expert judgments to capture dimensions that purely computational measures miss: linguistic nuance, cultural sensitivity, and instructional appropriateness. While models demonstrate impressive surface-level correction abilities, their explanations often lack the terminological and domain knowledge that effective language teaching requires, suggesting that current enthusiasm for AI-assisted language learning may be outpacing our understanding of these systems' actual pedagogical competence.

## Introduction

Large language models (LLMs) nowadays occupy several roles in language education: they correct grammatical errors, generate metalinguistic explanations, and approximate the functions of a human tutor (cf. Ali et al., 2024; Lo et al., 2024). At the same time, students increasingly use LLMs in their academic work to solve or clarify difficult topics, find study materials, do research, paraphrase, write reports and drafts, translate, or get instant feedback on their own work (Črček & Patekar, 2023; Yuan et al., 2024). Meanwhile, institutional bodies eagerly promote classroom adoption (The HG Foundation, 2025; Valchanov, 2024), and the empirical literature catalogues what these systems can do and where they fail. In contrast, Ye et al. (2025) argue that LLMs can serve as effective tutors in English education, assigning them roles from content generation to learner assessment to personalized instruction. The commercial sector has already acted on claims of this kind: between 2023 and 2025, Duolingo replaced over 100 contract writers and translators with AI-generated content (King, 2025), and users and linguists reported subsequent declines in cultural nuance and pedagogical accuracy, showcasing that the question that language pedagogy treats as settled, the technology sector treats as answered in the opposite direction. Namely, occupying these roles presupposes the ability to diagnose morphosyntactic errors, calibrate explanations to a learner's proficiency, and provide feedback that is both linguistically accurate and pedagogically well-timed. Whether current models actually possess these capacities in any robust sense remains an open question, one that adoption has outrun and left unanswered.

To test how effective current large language models are as teaching tools for teachers and learning tools for students, we compare three state-of-the-art model families (GPT, Claude, and Gemini) on their ability to identify, correct, and explain common learner errors in English. The study uses a small-scale dataset of twenty sentences, chosen based on empirical common learners' mistakes in the ESP (English for Specific Purposes) class, designed to represent frequent grammatical and ESP-related problems. ESP was chosen because of its unique position in English teaching, focusing on specific professional and occupational vocabulary and skills taught to students at the academic level, going beyond the generally taught concepts of traditional General English.  ESP falls within the broader field of ESL (English as a Second Language)[1] and it is oriented toward learners' academic or professional communicative needs, especially regarding practical outcomes (Dudley-Evans & St John, 1998; Anthony, 2018).

Drawing on established frameworks in error analysis (Corder, 1967), corrective feedback (Lyster & Ranta, 1997), and task-based language teaching (Ellis, 2003), we will analyze whether model-generated feedback can support the diagnostic and helper functions that effective language instruction usually requires. We will also observe the effects of prompt engineering, generation parameter variation, and retrieval-augmented generation (RAG) on output quality, pedagogical clarity, and response consistency. Since teachers and students typically interact with these systems through fixed consumer interfaces like ChatGPT rather than adjusting sampling parameters directly, the results are intended to also inform the design of specialized educational tools and deployment practices. RAG pipelines will be included to test whether external instructional materials improve feedback quality by supplementing the models' general training knowledge.

---

[1] In this paper, ESL and ESP are used interchangeably where the discussion concerns underlying language-learning processes. ESP is used where the reference is specifically to discipline-oriented instruction.

Automated metrics such as GLEU scores for fluency or BERTScore for semantic similarity cannot handle dimensions of pedagogical adequacy in a satisfactory way. These include linguistic nuance, cultural sensitivity, and instructional appropriateness, especially in ESP, which all require expert human judgment. The resulting dual-evaluation design is motivated by hallucinations – the tendency of language models to produce outputs that are confidently stated and factually wrong (Ji et al., 2023; Dziri et al., 2022). Šekrst (2025) argues that these hallucinations mirror cognitive errors and raise questions about the trustworthiness of AI-generated knowledge in domains where accuracy is non-negotiable. In educational contexts, a learner who receives a fluent, authoritative explanation of a grammatical or usage rule that happens to be incorrect is worse off than one who receives no explanation at all, because the error arrives with precisely the features that make it resistant to detection. The goal here was to show that, despite the success of LLMs as both teaching and learning tools, a successful implementation still requires careful human expertise and precision, in order not to impede the process it should support.

## Teaching with large language models

The empirical literature on AI in education has grown fast enough to sustain systematic reviews of its own. Ali et al. (2024) surveyed 112 articles and reported that ChatGPT can increase student engagement and accessibility. In language-specific research, Lo et al. (2024) reviewed empirical work on ChatGPT in ESL and EFL contexts (English as a Second and as a Foreign Language, respectively) and found use of AI for personalized learning, interactive practice, and dynamic feedback. However, they also found hallucinated explanations, uneven cultural sensitivity, and unresolved academic integrity pressures. Unfortunately, the interaction between the benefits and the risks has received little analytic attention: usually, the two sets of findings are catalogued side by side, as though documenting each were the same as understanding how they constrain one another.

A case study among Indonesian language educators found that 87% favored increased AI integration, citing expected improvements in vocabulary, grammar, and communication skills (Widianingtyas, Mukti & Silalahi, 2023). Oster, Henriksen, and Mishra (2024) report comparable results among teachers using ChatGPT for lesson planning, feedback generation, and assessment creation, where efficiency gains are acknowledged, and so are unresolved concerns about academic integrity, data privacy, and output accuracy. Both studies agree that responsible use requires structured educator training. However, some researchers are more cautious: Gundu and Chibaya (2024) offer targeted recommendations for extracting value from technology while being aware of its risks. Similarly, Hellen et al. (2024) argue that productive use in higher education requires competence in prompt engineering and responsible integration frameworks, especially in developing countries. In these studies, the enthusiasm for what technology can do coexists with uncertainty about the conditions under which it can do it *well*. Jeon and Lee (2023) identified four complementary ChatGPT roles in language instruction (interlocutor, content provider, teaching assistant, evaluator) and found that productive use depends on teachers curating AI outputs and guiding student interaction, a conclusion experiments will empirically substantiate.

## Technical challenges

The capabilities of large language models trace back to the transformer architecture (Vaswani et al., 2017). Models trained on massive text corpora acquire statistical regularities in language use and generate outputs by sampling from probability distributions over possible next tokens.[2] The model has no access to a fact-checking subroutine, and no internal distinction between what is *true* and what is merely *probable* given the training distribution. The result is that grammatically well-formed, confidently phrased outputs and factually incorrect, internally inconsistent, or wholly fabricated outputs are produced by exactly the same process the correct ones are produced as well (Ji et al., 2023).

The generation process is governed by parameter settings that control how the model samples from its probability distribution over possible next tokens. *Temperature* scales the distribution: low values concentrate probability mass on the highest-ranked tokens, producing predictable outputs, while high values flatten it, letting lower-ranked tokens through, which increases lexical diversity at the cost of coherence. Top-$k$ sampling restricts selection to the $k$ most probable tokens, and top-$p$ sampling includes tokens in descending probability order until their cumulative probability reaches a threshold $p$, so the candidate set expands or contracts depending on how concentrated the distribution is at each step.

[2] A token is the fundamental unit of text that a language model operates over. Most contemporary LLMs use subword tokenization, e.g., a word like *unforgettable* will be split into *un|forgett|able*, allowing the learning of various affixes and roots.

Our automated evaluation framework combines two automated metrics chosen for complementary reasons. GLEU (Napoles et al., 2015) is a sentence-level metric developed specifically for grammatical error correction that assesses how well generated corrections conform to natural language usage, without requiring reference texts. BERTScore (Zhang et al., 2020) evaluates semantic similarity between generated and reference texts using contextual embeddings from pre-trained language models. Both metrics are important for grammatical or lexical error correction in language teaching, where two explanations can convey the same meaning in entirely different words.

However, there are dimensions of pedagogical adequacy that GLEU and BERTScore are structurally incapable of measuring: whether a metalinguistic explanation is accurate, whether pedagogical scaffolding is appropriate to the learner's level, and whether the output is sensitive to cultural or professional and occupational context. Since LLM training data is known to carry systematic biases – which can produce outputs that reinforce stereotypes or fail to account for the linguistic and cultural diversity of actual learner populations – human judgment remains essential (Shah & Sureja, 2025).

## Language mistakes and teaching to correct them

Language acquisition necessarily involves producing errors and learning from them. This has been studied in applied linguistics at least since Corder (1967), who argued that learner errors are not annoying by-products of the learning process but evidence of the system the learner is building: an essential mechanism through which acquisition takes place. Superficially correct forms cannot be taken as evidence that the learner has internalized the corresponding linguistic system, as such forms may reflect imitation rather than genuine rule formation (Corder, 1967, p. 168). When learners let LLMs do the writing for them, they can produce polished output without actually testing their own hypotheses about the language. Any evaluation of LLMs in teaching needs to consider the diagnostic work that language instructors do by default.

As Lyster and Ranta (1997) demonstrate, *corrective feedback,* which can involve questions, metalinguistic comments, strategic pauses, repetition of errors or reformulations, but especially through elicitation, is most likely to engage learners when it leads to uptake through negotiation of form, meaning when teachers assist learners in reformulating an erroneous utterance without immediately supplying the correct form (Lyster & Ranta, 1997; Ellis, 2003, p. 80). In these cases, learners must draw on their own linguistic resources, a process Swain (2005) considers central to noticing gaps between what they intend to say and the linguistic means available to them. By contrast, the feedback typically provided by large language models tends to take the form of immediate reformulation, often without elicitation, prompting learner repair or offering metalinguistic explanation. This is the reason why we chose ESP materials for testing, where learner errors reflect unfamiliarity with professional vocabulary and conventions as much as with general grammar.

In ESP settings today, learners routinely use LLMs as translation tools, drafting assistants, essay writers, and task-solvers for discipline-specific assignments (Črček & Patekar, 2023; Cotton et al., 2023; Yuan et al., 2024). Many skip independent linguistic production entirely, prompting the model to generate target-like academic English for them. Studies confirm that university students in ESL and ESP contexts regularly use ChatGPT during drafting and revision, though interaction logs sometimes show learners requesting revisions, clarification, and evaluation of phrasing before submission (Han et al., 2023; Han et al., 2024).

From a task-based perspective, second language acquisition is a developmental process in which learners construct and progressively restructure a series of interlanguage systems over time rather than just accumulating discrete linguistic items (cf. Ellis, 2003). Learners pass through transitional stages of language acquisition, often over extended periods of exposure and use, before arriving at target-like forms. Learners may also produce target-like terminology in task-specific contexts without fully integrating those forms into their developing system, and this restructuring occurs through use and feedback. Instructors are uniquely positioned to do such tasks effectively since it involves active communication with the learner and prior knowledge of instruction dynamics. We do note that the models in this study were tested without learner-specific context, a constraint that a human instructor taking over a new class would also face. However, the domain-aware prompt condition provided the models with explicit information about the discipline, topic, and expected register, and this alone was sufficient to transform terminology detection from 21% to 89%.

Furthermore, error analysis has long recognized that the absence of errors in learner production does not reliably indicate target-like competence. Schachter (1974) showed that learners may simply avoid structures they find difficult, producing accurate but oversimplified output. If one only counts errors in what learners produce, it is easy to miss the possibility that learners are avoiding structures they find difficult. LLM-mediated writing creates a similar problem: when learners delegate production to a model, they can bypass forms that *exceed their current interlanguage* system entirely. The resulting text may be grammatically and lexically accurate at the surface level while the learner's underlying competence remains unchanged. In such cases, neither the student nor the teacher notices the discrepancy between what the learner can and should produce and what the target language requires, so there is no incentive for the learner to modify their response. The OECD Digital Education Outlook (OECD, 2026) reports convergent findings at scale: students using general-purpose GenAI tools produced higher-quality responses but performed worse on subsequent examinations, a pattern the report attributes to reduced cognitive effort during the AI-assisted task.

Recent work on interactional feedback has shown how difficult it is to interpret a learner's immediate repair after corrective feedback as genuine evidence of acquisition (Nassaji, 2020). Although modified output or repair may indicate that learners have noticed the feedback in some way, such responses may also reflect mechanical repetition. In ESP contexts, lexical development may involve the gradual strengthening of associative links between words, grammatical forms, and meanings through repeated exposure to co-occurring linguistic elements (Lightbown and Spada, 2006). As learners encounter discipline-specific terminology in meaningful contexts over time, connections between lexical items and their grammatical or semantic environments become increasingly strong. When an LLM supplies the target terminology directly, however, learners may produce the appropriate form without engaging in the repeated exposure and associative processes through which such lexical networks are typically developed.

## Case study

### *Methodology*

Our case-study evaluation proceeds in two stages. First, standard large language models (GPT, Claude, and Gemini) are tested on their ability to correct grammatical and terminological errors and to provide explanations using prompts that simulate real-world ESP/ESL learning scenarios. These prompts were constructed from common learner errors identified in homework tasks submitted by undergraduate students of biotechnical sciences through the e-learning platform as part of their regular ESP coursework. With the agreement of human ESP evaluators, these errors were iteratively used to construct the 20 learner sentences submitted to the models for correction. For example, students produced *human nutrition* where the discipline expects *diet*, and *heart and blood vessel diseases* where biotechnical English uses *cardiovascular diseases*. The models were asked to generate corrections with step-by-step feedback. Correction quality was assessed using GLEU and BERTScore, while pedagogical clarity and relevance were evaluated by ESP instructors.

Second, a GPT-5 model was tasked with generating assessment materials based on standard instructional prompts. In ESP instruction, lesson tests both assess retention and create opportunities for learners to process professional terminology under constraints that may promote acquisition. An LLM-generated test may therefore fulfil the first function while permitting success through shallow recognition of recently introduced forms. To examine this possibility, a GPT-5-generated test was compared with an instructor-designed assessment developed for the same lesson. The instructor-designed test required lexical reformulation and conceptual integration beyond recognition of recently introduced terms. Both assessments were voluntary, ungraded self-assessment tasks completed after the same instructional session.

Student performance data and engagement metrics from both task types were collected through the e-learning platform, including completion time, number of attempts, and final score. A total of 55 student attempts were recorded for the GPT-5-generated Lycopene revision task and 43 for the instructor-designed version. Each recorded attempt corresponds to a unique student, as the platform permitted multiple attempts but only graded the first attempt per participant. As participation was voluntary, the two cohorts are not identical; however, both tasks were completed by students enrolled in the same ESP course following the same instructional session. The minimum passing threshold for both assessments was set at 8 out of 15 points, allowing comparison of learner interaction across task formats, with time-on-task serving as a proxy for processing depth.

***Language errors***

To assess the reliability and methodological quality of LLM-generated corrective feedback under varying model generation parameters, we evaluated model outputs on twenty English sentences (see Appendix A) containing errors typical of ESP learner production, distributed across four categories: subject-verb agreement (sentences 1–5), article usage (sentences 6–10), mixed grammatical errors involving tense, word form, preposition, and syntactic structure (sentences 11–15), and ESP-specific terminology and register (sentences 16–20). The first three categories test standard grammatical error correction, a prerequisite for any system used in a feedback role.

The fourth part is qualitatively different: sentences 16–20 are grammatically well-formed but *terminologically or stylistically inappropriate* for their target discourse community. They contain lay paraphrases of professional biotechnical terminology drawn from instructional materials on lycopene, requiring the model to recognize register-level inadequacy and propose discipline-appropriate alternatives. The fifteen grammar sentences (1–15) contained a total of 39 distinct error tokens, each scored independently for correct identification and repair; individual sentences contained between one and four errors.

Grammar and overall linguistic accuracy were evaluated by two independent raters with postgraduate training in applied linguistics and experience teaching English for Specific Purposes at the university level. ESP terminology and register adequacy were evaluated by one domain expert in biotechnical English and one ESP instructor with experience teaching biotechnology students. Raters independently reviewed all model outputs and compared them against gold-standard reference corrections. Disagreements were resolved through discussion, with consensus reached in all cases. Agreement prior to discussion exceeded 90% across both grammatical and terminological judgments, indicating high inter-rater reliability for the evaluation procedure.

For each model, a standardized prompt (see Appendix B) instructed the system to identify all errors, provide a corrected version, and explain each correction. This prompt design reflects the tripartite structure of corrective feedback described by Lyster and Ranta (1997): error occurs, feedback is provided, uptake takes place. We have, however, combined explicit correction and metalinguistic feedback into multiple feedback (Lyster & Ranta, 1997, p. 48) due to the possibilities that LLMs offer. Generation parameters were varied systematically across configurations for each model to assess output consistency; the full parameter specifications are reported in Appendix D.

### *Findings*

The results split cleanly along a line that is relevant to ESP instruction. Table 1 summarizes performance on the two evaluation dimensions.

| Model | Architecture | Grammar (S1–15) | ESP (S16–20) | Average output tokens | Average latency |
|---|---|---|---|---|---|
| GPT-4o | Standard | 97% (38/39) | 2/19 (11%) | 274 | 5.0s |
| GPT-5 | Reasoning | 100% (39/39) | 7/15 (47%)[3] | 1,993 | 30.4s |
| Claude Opus 4.6 | Standard | 97% (38/39) | 15/19 (79%) | 615 | 14.7s |
| Claude Sonnet 4.6 | Standard | 97% (38/39) | 5/19 (26%) | 515 | 11.3s |

[3] GPT-5 ESP score is based on 15 evaluable terms; sentence 19 produced no output under the allocated token budget (see Appendix D.3).

| Gemini 3 Pro | Reasoning | 97% (38/39) | 15/19 (79%) | 553 | 20.1s |
|---|---|---|---|---|---|
| Gemini 3 Flash | Standard | 97% (38/39) | 11/19 (58%) | 603 | 11.3s |

**Table 1: Models tested and their averaged results**

For the error types examined in this study, grammatical correction appears highly reliable, as confirmed by external and internal evaluators. Every model achieved 97% accuracy or higher on sentences 1–15, and the residual misses were possible alternative corrections. GPT-4o's parameter sweep confirmed that identical corrections appeared across all configurations for nearly every grammatical item. Subject-verb agreement, article usage, tense errors, and common morphosyntactic patterns occupy high-probability regions of these models' output distributions, and the corrections they produce are reliable. However, the grammatical items examined here are common sentence-level morphosyntactic errors typical of ESP contexts, and performance on these items does not imply consistent reliability across all aspects of learner language. Large learner corpora include more complex issues such as pragmatic errors and patterns influenced by different first-language backgrounds, and performance may also vary in low-resource contexts and less-represented languages.

ESP terminology, on the other hand, tells a different story. Performance ranged from 11% (GPT-4o) to 79% (Claude Opus and Gemini Pro), a sevenfold difference on the same task with the same prompt. Claude Opus detected three times as many ESP terms as Claude Sonnet (15 versus 5); Gemini Pro outperformed Gemini Flash by a margin (15 versus 11). Both pairs share architectural families and training data lineages, yet the larger model in each case substantially outperformed the smaller one on terminology while producing nearly identical grammar scores. This shows that grammatical correction and domain-specific terminology detection rely on different abilities: the first works well across model sizes, while the second drops off in smaller models that lack the needed capability.

These numbers are informative, but what a learner or instructor actually encounters is a single piece of feedback. Consider sentence 17: *The red color of tomatoes comes from a fat-liking colored substance that belongs to the family of plant colors and helps protect against heart and blood vessel diseases.* This sentence is grammatically correct, however, an ESP instructor would identify four terminological problems: *fat-liking* should be *lipophilic* (or *fat-soluble*), *colored substance* should be *pigment*, *family of plant colors* should be *carotenoid family*, and *heart and blood vessel diseases* should be *cardiovascular diseases*. The sentence communicates the right meaning, but in the wrong register. Even Claude Sonnet 4.6 failed to recognize register appropriateness in discipline-specific writing.

GPT-4o, under default settings, returned the sentence with minor stylistic adjustments and no terminological corrections. The lay paraphrases were treated as acceptable. A student receiving this feedback would have no reason to suspect that fat-liking colored substance is inadequate for biotechnical writing. Claude Opus 4.6 replaced *fat-liking* with *fat-soluble*, added the gloss *lipophilic*, identified *colored substance* as vague, proposed *carotenoid pigment*, and corrected *heart and blood vessel diseases* to *cardiovascular diseases*, explaining why each lay term is inappropriate in scientific discourse. GPT-5, the reasoning model, produced a comparable correction, noting that pigments are, by definition, colored and that the lay phrasing is redundant and imprecise. Both responses approximate what an ESP instructor would provide.

The grammatical items produced a different kind of instructive variation. Sentence 4 (*Neither the results nor the methodology suggests that the hypothesis are correct*) involves a genuine area of prescriptive disagreement: whether the verb should agree with the nearer noun (proximity agreement, yielding suggests) or with the coordinated subject (notional agreement, yielding suggest). GPT-4o silently switched between the two forms across parameter settings, providing whichever was sampled without acknowledging the ambiguity. GPT-5 selected and explained the proximity agreement rule, noting that reversing the noun order would allow the plural verb. Claude Opus 4.6 showed self-correction in its output, considering both options before choosing a form that explains the competing rules. From Lyster and Ranta's (1997) perspective, the latter two responses turn grammatical ambiguity into a learning opportunity, as metalinguistic feedback and repetition are more likely to elicit uptake, whereas the first does not.

### *Prompt specificity and retrieval augmentation*

The most consequential finding of the evaluation did not involve model selection. To resolve the terminological adequacy issues, we compared three prompting conditions on the five ESP sentences using GPT-4o and GPT-5 (see Appendix C for full prompts). We chose GPT models since they are often exactly the ones students and teachers use via ChatGPT in everyday practice. The first was the generic prompt (the baseline for this comparison). The second is the professional/occupational context: an ESP course on food science, following lessons on lycopene, with attention to terminological accuracy and register. The third one augmented this domain-aware prompt with retrieval from course-relevant textbooks and lecture materials.

Under the generic prompt, GPT-4o detected 4 of 18 target terms, and under the domain-aware prompt, the same model detected 17 of 18. This fourfold increase was produced by a single modification: *naming the discipline, topic, and expected register*, as agreed by human evaluators. Terms entirely absent under the generic prompt, including *bioactive compounds*, *lipophilic*, *carotenoid*, *cardiovascular*, *lycopene*, and *scavenge*, were activated reliably once the prompt specified the food science domain. The domain-aware prompt that raised GPT-4o from 4/18 to 17/18 was written by an ESP instructor who knew the discipline, the topic, and the target register: prompt specificity, in this case, is teacher expertise encoded as text. Teacher knowledge is the variable that determines whether LLM feedback is pedagogically adequate, and the results show this holds regardless of whether that knowledge reaches the learner through classroom instruction or through prompt design.

The effect of retrieval augmentation – adding textual chunks of course-adjacent material – was configuration-dependent. Under the original settings (800-word chunks, $k$=10), RAG achieved perfect term detection (18/18) on GPT-5, the only condition in the study to reach this ceiling, exceeding the domain-aware prompt alone (17/18). However, a follow-up ablation varying chunk size (200 and 800 words) and retrieval depth ($k$=5 and $k$=10) showed that three of the four alternative configurations scored 15/18, underperforming the domain-aware prompt without retrieval. For GPT-4o, the RAG condition yielded the same 17 of 18 terms as the domain-aware prompt. Smaller chunks (200 words) performed consistently worse regardless of retrieval depth, suggesting that the retrieved fragments lacked sufficient surrounding context to activate domain terminology.

***Error correction discussion***

Findings related to pedagogical questions raised in the text, along with practical implications for ESP instruction, can be stated concisely. First, grammatical error correction can be reliably delegated to any current-generation language model, and the model selection is largely irrelevant for this task. An ESP instructor or a student using any of the six models tested would receive accurate grammatical corrections on learner's writing of the kind represented in this study.

Second, feedback on ESP terminology and register cannot be delegated without careful attention to prompt design and model capability. Although ESP terminology is present in the models' training data, it requires contextual activation. Generic prompts do not reliably elicit domain-specific vocabulary, whereas prompts that specify the discipline, instructional topic, and expected register do. Retrieval augmentation may help when well-tuned, but when poorly-tuned, it can introduce noise, reduce detection below prompt-only performance, or shift output toward phrasing that is too technical for classroom use. In practice, however, most teachers and students access these models through consumer interfaces such as ChatGPT, where retrieval cannot be configured and prompt design remains the only available lever. Improving prompt design therefore proved more effective than model selection, parameter tuning, or retrieval infrastructure, and it could be applied in a standard non-specialist usage of LLM tools.

A student submitting a terminologically inadequate sentence to GPT-4o under default conditions may be told the sentence is acceptable, while the same sentence submitted to Claude Opus 4.6 would receive the domain-specific corrections an instructor would provide. In Corder's (1967) terms, a model that treats lay terminology as adequate fails to create the conditions under which the learner might notice the gap between interlanguage and target register. Lyster and Ranta (1997) and Mackey (2006) treat such errors as the raw material of corrective feedback in meaning-focused instruction. When AI-mediated production removes them before the instructor sees the text, the occasion for metalinguistic explanation during class discussion is lost.

A related problem appeared when the twenty evaluation sentences (see Appendix A) were submitted to ChatGPT, used here as a proxy for the consumer-facing deployment of the GPT-5.2 model through which students typically interact with the system, *without prompt modification*, approximating how students actually use the tool. The model returned corrected versions *with no explanation of what was wrong or why*. This behavior resembles what Lyster and Ranta (1997) classify as recasts: reformulations that supply the target form without drawing the learner's attention to the specific feature that was corrected. Recasts are the least likely type of feedback to elicit learner uptake or self-repair in communicative classrooms. Effective pedagogical feedback does something recasts do not: it identifies the point of divergence, explains why the learner's form is inadequate, and creates conditions under which the learner might notice the discrepancy between their production and the target (Schmidt, 1990).

As Swain (2005) argues, without opportunities to test hypotheses about form-meaning relationships, learners have no mechanism for restructuring their interlanguage. An LLM that silently replaces *heart and blood vessel diseases* with *cardiovascular diseases* has solved the learner's immediate communicative problem, but it has not created any condition under which the learner might understand why the first formulation is inadequate in biotechnical discourse, or remember to produce the second one next time. An instructor, situated within the learner's history, the task demands, and the conventions of the target discipline, can do both (Ellis, 2003; Long, 2015; Lyster & Ranta, 1997).

Finally, the difference between a generic prompt and a domain-aware prompt (4/19 to 17/19 on GPT-4o) exceeded the difference between any two models in the main evaluation. The instructor's role, in this configuration, moves from providing grammatical corrections to specifying the domain context that activates the model's latent terminological knowledge and to evaluating whether the resulting feedback meets the register expectations of the target discourse community. This is a supervisory role that requires the domain expertise and methodological judgment that the models themselves still lack.

The automated metrics confirm this pattern while exposing a methodological limitation. On grammatical sentences, GLEU and BERTScore converged at the ceiling for every model, reflecting the uniformity of the corrections. On ESP sentences, however, BERTScore remained high (0.92–0.96), indicating semantic adequacy, while GLEU dropped sharply (0.25–0.50), reflecting lexical distance from the reference terminology. Models proposing scientifically valid but lexically divergent terms (e.g., *fat-soluble* instead of *lipophilic*) received lower GLEU scores despite adequate meaning.[4] Developing metrics sensitive to register, domain terminology, and scaffolding quality is an open problem that our results make concrete.

Retrieval augmentation did not provide additional benefit when the prompt was already domain-specified, though its effects were configuration-sensitive: one configuration (800-word chunks, $k$=10) achieved perfect detection on GPT-5, while most underperformed the domain-aware prompt alone. Its pedagogical value is therefore conditional on retrieval design, which requires both technical and domain expertise.

These findings should be interpreted with appropriate caution. The evaluation involved twenty sentences, five of which tested ESP terminology, across a single disciplinary domain. The models were evaluated in their default, publicly available configurations without fine-tuning for methodological or domain-specific tasks. A larger evaluation spanning multiple ESP domains, more complex error types, and fine-tuned model variants might produce a different hierarchy or narrow the gaps reported here. The magnitude of the prompt-specificity effect, however, is large enough to be pedagogically meaningful even at this scale, and the consistency of the grammar/terminology dissociation across all six models and three providers suggests it reflects a structural property of current language models.

***Lesson tests discussion***

The error correction evaluation examined whether LLMs can respond adequately to learner-produced text. A separate question is whether they can generate adequate educational materials. The second component of the case study compared two lesson tests on the same topic (lycopene) administered to second-year biotechnology students through the e-learning platform. For example, a student who scores 14/15 in forty seconds and a student who scores 14/15 in six minutes have produced the same outcome through qualitatively different cognitive operations. The former is consistent with pattern recognition, and the latter suggests effortful integration of form and meaning. Nassaji (2020) has argued that process-oriented indicators of engagement capture dimensions of learning that post-task accuracy misses, and this argument applies with particular force in ESP contexts, where successful performance requires mapping professional terminology onto conceptual structures from the learner's field.

[4] One caveat is necessary: GPT-4o's ESP GLEU (0.476) fell only marginally below Opus's (0.501), despite detecting far fewer terms (2/19 versus 15/19). This compression occurs because GLEU evaluates the full corrected sentence, and high overlap on non-terminological portions masks substantial terminological divergence. Automated metrics should therefore be interpreted alongside targeted term-matching analyses when evaluating ESP correction quality.

The GPT-5-generated test required recognition of recently encountered terms in familiar contexts, whereas the instructor-designed test required integration. Several items asked students to select English equivalents of the terms in their mother tongue within a domain-specific frame, to discriminate between lexically similar alternatives, or to apply conceptual knowledge to terminological choice (for instance, identifying *lipophilic* as the relevant property in a question about membrane transport). Both tests used a multiple-choice format, but the cognitive demands were not equivalent: selecting lipophilic from a set of alternatives because one remembers seeing the word in a slide is a different operation from selecting it because one understands that fat-solubility determines a compound's capacity to cross a lipid bilayer. The first is recall, and the second requires the learner to evaluate form-meaning mappings and suppress plausible but incorrect alternatives. Schmidt (1990) and Swain (2005) treat this kind of effortful evaluation as the mechanism through which noticing and hypothesis testing drive interlanguage restructuring.

Gass and Selinker (2008) make a similar argument: acquisition proceeds when learners are forced to reconcile new forms with existing conceptual structures, and task formats that permit shallow recognition provide fewer occasions for this reconciliation.

Performance on the ChatGPT test was uniformly high (M = 13.96/15). Ninety-four percent scored 12 or above, half achieved a perfect score, and no student fell below the passing threshold, suggesting the task was solvable through recognition of recently encountered terms. However, the instructor-designed test produced comparable accuracy scores (13.39/15), but completion times diverged. Students spent an average of 5 minutes and 28 seconds on the instructor-designed test (M=328s), while the ChatGPT test was completed substantially faster (M=264s, 4 minutes 24 seconds). The accuracy match combined with the time difference tells us that both groups performed well, but one group took significantly longer to do so, roughly 24% longer. Students' response time in this case cannot be taken as direct evidence of acquisition, but can be an indirect measure of processing effort. Namely, in ESP contexts, where correct answers require mapping terminology onto discipline-related concepts, longer processing times are consistent with the effortful integration that Schmidt (1990) and Swain (2005) associate with noticing and hypothesis testing, with deeper processing involving a more elaborate effort and stronger memory traces (Swain, 2005). Tasks solvable through rapid pattern recognition, by contrast, provide fewer occasions for the error-driven restructuring that Corder (1967) identifies as the mechanism of interlanguage development. Furthermore, testing is an effective way to improve the retention of lecture material (Butler & Roediger, 2007), while introducing challenges or difficulties for the learner fosters long-term goals of training (Metcalfe & Shimamura, 1994), as more complex retrieval processes during testing are more supportive of long-term retention (Bjork, 1975). Thus, more difficult tests used as learning tools after a lesson are more beneficial for students as they help them retain the target material better.

## Transparency and accountability

The findings reported above have implications that extend beyond the question of which model performs best, raising an important question: under what conditions can a learner or instructor know *whether the feedback they are receiving is adequate*? The answer, at present, is that they cannot. These silent failures are structurally invisible to the students who encounter them, and indistinguishable from success without the very domain knowledge the tool is meant to provide, leading us to an accountability problem. In conventional instruction, the feedback chain is traceable: a student can ask the instructor why a correction was made, and the instructor can explain the reasoning, cite the relevant convention, and acknowledge uncertainty where it exists. The exchange is situated within a relationship in which both parties understand who is responsible for the quality of the explanation. With LLM-generated feedback, this chain breaks since the model does not know why it produced one correction rather than another, and the student often fails to, or refuses to, interrogate the reasoning. The instructor, if one is involved at all, may never see the output.

The difference between adequate and inadequate ESP feedback turned on a single design decision (whether the prompt named the discipline and topic) that is invisible to the learner and, in most deployment contexts, to the instructor as well. A student using ChatGPT through its consumer interface usually does not change the system prompt telling the model how to behave and has no way to determine whether the feedback reflects a configuration that activates domain-specific terminology or one that does not.[5] Two students submitting the same sentence to the same model on the same day could receive categorically different feedback.

Institutional responses to date have focused on academic integrity (cf. Balalle & Pannilage, 2025; Chan, 2025): whether students are submitting AI-generated text as their own. Of course, this is a real concern, but it addresses the wrong end of the problem. The more consequential risk is that they use LLMs to *learn*, and that the learning is corrupted by feedback they have no means of evaluating. A student who copies an essay has at least circumvented the pedagogical process, but a student who submits their own writing, receives confident but terminologically vacuous feedback and internalizes it as adequate has been actively misdirected.

Responsible deployment requires, at a minimum, three things. First, we need transparency about what the model can and cannot evaluate. That is, an LLM used in an ESP context should state that it has not been configured for domain-specific terminology unless explicitly prompted to do so, and current models do not do this by default and without technical engineering.

Second, traceability of the feedback chain is important as well since instructors who incorporate LLM-generated feedback into their teaching need access to the prompts, parameters, and model versions that produced it. Without this, they cannot assess whether the output meets the standards of their discipline.

[5] An important note is that interfaces like ChatGPT do not allow tweaking the specific model parameters as we did in our study.

Third, institutional policies that address feedback quality alongside academic integrity are of equal importance as well. If universities are going to endorse or tolerate the use of LLMs in language learning, they have an obligation to ensure that the feedback students receive from those tools is subject to the same quality standards they apply to human instruction (cf. UNESCO, 2025).

## Concluding remarks

The question posed by this paper's title has a probably expected but still straightforward answer: no. Current large language models reliably correct English grammar in standard cases. Every model tested achieved 97% accuracy or higher on standard morphosyntactic errors, and this performance was robust to parameter variation, model scale, and architectural differences across three providers. An ESP instructor can probably delegate grammatical error correction to any current-generation LLM without meaningful loss of accuracy: this is a genuine capability, expected since the models were mostly trained on English language data, and it would be insincere to deny it.

What the models cannot do, without substantial human input, is the rest of the job. ESP terminology detection ranged from 11% to 79% across models, a sevenfold spread that disappeared almost entirely when the prompt named the discipline, topic, and expected register. The terminology was in the training data all along, but it required activation, and activation required an instructor *who knew what to ask for*. Analogously, both the ChatGPT-generated and instructor-designed tests produced high accuracy scores, but the ChatGPT test was completed in a fraction of the time, with half the cohort achieving perfect scores (in a way, echoing Schachter (1974)). Tasks that permit success through recognition of recently encountered forms provide fewer occasions for the effortful integration that drives interlanguage restructuring (Schmidt, 1990; Swain, 2005; Corder, 1967).

So, *can LLMs replace language teachers in any way*? The capacities that LLMs possess (grammatical correction, surface-level reformulation, and generation of recognition-based test items) are the ones that are easiest to automate and least central to language acquisition. The capacities they lack (domain-sensitive terminology detection without explicit prompting, generation of tasks that require deep processing, production of feedback that creates conditions for noticing or elicitation) are the ones that define what a language instructor does. This is similar to modern software developers: when prompted well, AI can create good, robust, and secure code, but only if led by a professional using natural language as a set of professional instructions, drawing from their own knowledge.

The technology is useful in proportion to the expertise of the person directing it. An instructor who understands the target discourse, who can specify the domain context that activates the model's latent knowledge, and who can evaluate whether the resulting output meets the register expectations of the discipline will find these tools genuinely productive. A student using ChatGPT with a generic prompt will receive grammatically correct, terminologically imprecise feedback that implicitly confirms that lay paraphrases are adequate for scientific writing. The distance between these two outcomes is the distance between a tool and a teacher.

## Appendix A: Evaluation sentences

Sentences 1–15 contain deliberate grammatical errors representative of those commonly produced by learners in ESP instructional settings. Sentences 16–20 are grammatically well-formed but terminologically and stylistically inappropriate for their target discourse community: they contain lay paraphrases of domain-specific biotechnical terminology drawn from instructional materials on lycopene.

*Category 1: Subject–verb agreement (sentences 1–5):*

1. The data shows that the reaction rate increase significantly at higher temperatures.
2. Each of the samples were tested for contamination before the analysis were completed.
3. The number of participants in the study are relatively small compared to previous research.
4. Neither the results nor the methodology suggests that the hypothesis are correct.
5. The committee have decided that the criteria for selection is too strict.

*Category 2: Article usage (sentences 6–10):*

6. Researcher conducted a experiment to determine the effect of UV light on a bacteria.
7. The lycopene is powerful antioxidant found in the tomatoes and other red fruits.
8. Patient was administered an dose of the medication before undergoing a MRI scan.
9. In conclusion, the further research is needed to understand role of the vitamin D in immune function.
10. She has degree in the biochemistry and works as researcher at university.

*Category 3: Mixed grammatical errors (sentences 11–15):*

11. The study was conduct by a team of researchers who has been working on this project since three years.
12. According to the results, the treatment effected positively on patient recovery and reduced the number of side affects.
13. The students which participated in the experiment was asked to complete a questionnaire about their experience with the learn materials.
14. It is important that the researcher considers all the possible variables that might to influence the outcome of the study.
15. The amount of informations gathered during the survey were insufficient to drawing any meaningful conclusions.

*Category 4: ESP-specific terminology and register (sentences 16–20):*

16. Tomatoes and tomato products have always held a special place in human nutrition due to their rich content of important substances that can prevent damage caused by harmful molecules.
17. The red color of tomatoes comes from a fat-liking colored substance that belongs to the family of plant colors and helps protect against heart and blood vessel diseases.
18. The substance that gives tomatoes their red color occurs in a crystal form and is not easily taken in by the body unless the tomatoes are broken down through heating.

19. Scientists have found that eating things made from tomatoes, like paste or sauce, with some fat helps the body to use the red substance better because it likes to join with fat-like materials.
20. The red plant color can hunt for dangerous atoms in the body, which may help lower the chance of getting bad cholesterol buildup in blood vessels.

## Appendix B: Standardized evaluation prompt

The following prompt was used across all models and parameter configurations. The system prompt established the evaluator role, while the user prompt presented each sentence and specified the required response structure.

System prompt:

> You are an experienced English language instructor specialising in English for Specific Purposes (ESP) and academic writing. Your task is to provide corrective feedback on learner writing.

User prompt:

> The following sentence was written by an English language learner. Identify all errors in the sentence, provide the corrected version, and explain each correction in a way that would help the learner understand and apply the underlying rule in future writing.
> Sentence: "{sentence}"
>
> Please structure your response as:
> 1. CORRECTED SENTENCE: [the corrected version]
> 2. ERRORS IDENTIFIED: [list each error]
> 3. EXPLANATIONS: [for each error, explain the rule and why the correction applies]

The tripartite response structure (correction, identification, explanation) reflects the components of effective corrective feedback described by Lyster and Ranta (1997). The placeholder {sentence} was replaced with each of the twenty evaluation sentences listed in Appendix A using a Python 3.14 script.

## Appendix C: Domain-aware and RAG evaluation prompts

The domain-aware prompt was designed to test whether naming the specific instructional context would activate terminology that the generic prompt failed to elicit. Three elements were specified: the discipline (e.g., food science), the topic (e.g., lycopene), and the expected register (e.g., academic/scientific) and these correspond to the contextual information an ESP instructor would bring to any feedback interaction. The RAG condition added a further layer by supplying textual chunks from course-relevant materials, testing whether retrieval from instructional sources could compensate for what the model's training data alone could not provide. Here, retrieval parameters were selected after preliminary testing indicated that smaller chunks, and lower retrieval depths produced incomplete terminological coverage (the ablation in Appendix D.5 confirms this: only the 800-word, k=10 configuration matched or exceeded the domain-aware baseline).

The following prompts were used for the ESP terminology evaluation described in Section *Prompt specificity and retrieval augmentation*. The baseline condition used the generic prompt from Appendix B. The domain-aware and RAG conditions used the modified prompts below.

Domain-aware system prompt (used in both domain-aware and RAG conditions):

> You are an experienced English language instructor specialising in English for Specific Purposes (ESP) and academic writing. Your task is to provide corrective feedback on learner writing, with particular attention to domain-specific terminology and scientific register.

RAG augmentation (appended to the domain-aware system prompt when retrieval was active):

> You have access to the following reference material from ESP course materials and textbooks on lycopene and food science. Use this material to identify cases where students have used lay or informal paraphrases instead of the correct domain-specific terminology.
>
> REFERENCE MATERIAL: [Retrieved chunks inserted here]
>
> When evaluating student writing, pay particular attention to:
> 1. Standard grammatical errors (agreement, articles, tense, etc.)
> 2. Terminological accuracy: students should use domain-specific terms, not lay paraphrases
> 3. Register appropriateness: academic/scientific register is expected.
> 4. If a lay term has a standard scientific equivalent shown in the reference material, flag it and provide the correct term.

For each sentence, the ten most semantically similar chunks were retrieved from the corpus using cosine similarity over embeddings generated by OpenAI's *text-embedding-3-small* model. The retrieval corpus comprised 390 chunks (800 words each, 200-word overlap) derived from four course-relevant lecture materials.

Domain-aware user prompt (replacing the generic user prompt from Appendix B):

> The following sentence was written by an English language learner in an ESP (English for Specific Purposes) course on food science, following lessons on lycopene. Identify all errors in the sentence, including any cases where the student has used informal or lay language instead of the correct domain-specific terminology, provide the corrected version, and explain each correction.
>
> Sentence: "{sentence}"
>
> Please structure your response as:
>
> 1. CORRECTED SENTENCE: [the corrected version using appropriate scientific terminology]
> 2. ERRORS IDENTIFIED: [list each error, including terminological/register issues]
> 3. EXPLANATIONS: [for each error, explain the rule and why the correction applies]

## Appendix D

Evaluation scripts, anonymized JSON outputs, e-learning platform (Merlin) interaction logs, and supporting documentation for the human evaluation procedure are available in the paper repository https://github.com/ksekrst/llm-esp. The repository also includes certification statements from the applied linguistics and biotechnology experts who participated in the human evaluation, documenting their qualifications and role in the assessment process.

### D.1 Model configurations

| Model | Provider | Architecture | Configurations | API calls | Max tokens |
|---|---|---|---|---|---|
| GPT-4o | OpenAI | Standard | 4 temp × 3 top-*p* | 240 | 1,500 |
| GPT-5 (GPT-5.2) | OpenAI | Reasoning | 1 (fixed) | 20 | 1,500 / 4,096 |
| Claude Opus 4.6 | Anthropic | Standard | 1 (temp 0.0) | 20 | 1,500 |
| Claude Sonnet 4.6 | Anthropic | Standard | 1 (temp 0.0) | 20 | 1,500 |
| Gemini 3 Pro | Google | Reasoning | 1 (deterministic) | 20 | 1,500 / 8,192 |
| Gemini 3 Flash | Google | Standard | 1 (deterministic) | 20 | 1,500 |

**Table 2: Model configurations**

Table 2 shows model configurations used. GPT-4o temperatures: 0.0, 0.3, 0.7, 1.0; top-*p:* 0.5, 0.9, 1.0. GPT-5 (GPT-5.2 used throughout the study) locks temperature and top-*p* at 1. Anthropic API does not permit simultaneous specification of temperature and top-*p*. Dual token values indicate initial and revised budgets for reasoning models. Models were tested with their default versions via API endpoints as *gpt-4o, gpt-5, claude-opus-4-6, claude-sonnet-4-6, gemini-3-pro-preview,* and *gemini-3-flash-preview* from Feb 1 to Feb 22, 2026, receiving negligible differences (no specific subversions pinned).

### D.2 Cross-model results

| Model | Grammar (S1-S15) | ESP terms (S16-20) | Average tokens | Average latency |
|---|---|---|---|---|

| Claude Opus 4.6 | 97% (38/39) | 15/19 (79%) | 615 | 14.7s |
|---|---|---|---|---|
| Gemini 3 Pro | 97% (38/39) | 15/19 (79%) | 553 | 20.1s |
| Gemini 3 Flash | 97% (38/39) | 11/19 (58%) | 603 | 11.3s |
| GPT-5 | 100% (39/39) | 7/15 (47%)* | 1,993 | 30.4s |
| Claude Sonnet 4.6 | 97% (38/39) | 5/19 (26%) | 515 | 11.3s |
| GPT-4o | 97% (38/39) | 2/19 (11%) | 274 | 5.0s |

**Table 3: Cross-model results**

Table 3 shows cross-model results ranked by ESP detection. *GPT-5 ESP score out of 15 evaluable terms (sentence 19 produced no output in the last recorded run). Residual grammar misses were defensible alternatives (sentence 6: “The researcher” for “A researcher”; sentence 12: restructuring around “affected”). GPT-5 token counts include hidden reasoning overhead.

**D.3 Reasoning Model Behavior**

At a 1,500-token budget, GPT-5 returned empty responses for 15 of 20 sentences (75%), consuming the full allocation on internal reasoning. At 4,096 tokens, 19 of 20 completed. The estimated proportion of tokens consumed by hidden reasoning ranged from 73% (sentence 7) to 91% (sentence 4), with ESP sentences averaging 87%. Gemini Pro at 1,500 tokens truncated 5 of 20 responses and leaked reasoning traces in two cases: sentence 1 surfaced “Teacher Persona):**

Tone: Encouraging, clear, authoritative but accessible"; sentence 4 surfaced "But 'methodology' is closer. So 'suggests' is correct." At 8,192 tokens, all 20 completed.

**D.4 GPT-4o parameter sensitivity**

For sentences 1–15, corrections were identical across all 12 temperature–top-*p* configurations. Sentence 4 ("neither…nor") was the sole exception: GPT-4o alternated between "suggest" and "suggests" in a 6/12 split with no systematic relationship to parameter settings. Average output length: 269 tokens at temperature 0.0, 278 at 1.0. On ESP sentences, parameter variation produced visible effects. Sentence 20 yielded eight distinct corrected versions across twelve runs; sentence 19 yielded six. Unique corrections increased monotonically with temperature for sentence 19 (one variant at 0.0, three at 1.0).

**D.5 Retrieval-augmented generation**

| Model | Generic prompt | Domain-aware prompt | Domain-aware prompt + RAG |
|---|---|---|---|
| GPT-4o | 4/18 | 17/18 | 17/18 |
| GPT-5 | — | 17/18 | 18/18 |

**Table 4: Retrieval-augmented generation**

Table 4 shows ESP term detection by prompting condition. Retrieval corpus: 390 chunks (~800 words, 200-word overlap) from Rao, Young, and Rao (2021), Pokorný, Yanishlieva, and Gordon (2001), and lecture materials. Embeddings: *text-embedding-3-small.* Retrieval: 10 chunks per sentence by cosine similarity (range 0.55–0.65). GPT-5 was not tested under the generic prompt.

A follow-up ablation on GPT-5 varied chunk size (200, 800 words) and retrieval depth ($k$=5, 10) under the domain-aware RAG condition with text-embedding-3-small. Only the 800-word, $k$=10 configuration matched or exceeded the domain-aware prompt baseline (18/18 vs. 17/18); configurations with 200-word chunks scored 15/18 regardless of $k$, and 800-word chunks at $k$=5 also scored 15/18.

**D.6 Automated metrics**

GLEU (Napoles et al., 2015) and BERTScore (Zhang et al., 2020) were computed for all 339 responses against gold-standard references that incorporated full target ESP terminology.

| Model | Gram. GLEU | Gram. BERTScore | ESP GLEU | ESP BERTScore |
|---|---|---|---|---|
| Claude Opus 4.6 | 0.980 | 0.999 | 0.501 | 0.96 |
| Gemini 3 Pro | 0.937 | 0.998 | 0.498 | 0.95 |
| GPT-5 | 0.946 | 0.998 | 0.494 | 0.95 |
| GPT-4o | 0.943 | 0.997 | 0.476 | 0.96 |
| Gemini 3 Flash | 0.969 | 0.999 | 0.473 | 0.95 |

| Claude Sonnet 4.6 | 0.953 | 0.998 | 0.254 | 0.92 |
|---|---|---|---|---|

**Table 5: GLEU and BERTScore metrics**

Table 5 shows automated GLEU and BERTScore metric results. Sonnet's low ESP GLEU scores align with sentences 16 and 18, where the model assessed the input as "grammatically correct and well-constructed" and proposed no corrections (GLEU 0.017 and 0.042; BERTScore 0.835). GPT-4o's ESP GLEU appears high relative to its term detection (2/19) because preserved original phrasing overlapped with the reference in non-terminological portions.

| Condition | 4o GLEU | 4o BERTScore | 5 GLEU | 5 BERTScore |
|---|---|---|---|---|
| Domain-aware | 0.576 | 0.967 | 0.192 | 0.917 |
| Domain + RAG | 0.560 | 0.968 | 0.165 | 0.916 |

**Table 6: RAG condition metrics, ESP sentences only**

Table 6 shows retrieval-augmented generation with attached course materials using ESP sentences from Category 4 (Appendix A) for evaluation. GPT-5's low GLEU despite strong term detection (17/18) reflects extensive sentence rewriting; BERTScore confirms semantic content is preserved.